%% file: paper.tex
\documentclass[]{ailab}

\usepackage{tcolorbox}
\usepackage{fontawesome}
\usepackage{hyperref}
\usepackage{wrapfig}
\usepackage{tabularx}
\usepackage{textcomp}
\usepackage{stfloats}
\usepackage{url}
\usepackage{verbatim}
\usepackage{titlesec}
\usepackage{adjustbox}
\usepackage{multirow}
\usepackage{pifont}
\usepackage{tikz}
\usepackage{comment}
\usepackage{amsmath,amssymb}
\usepackage{colortbl}
\usepackage{color}
\usepackage{booktabs}
\usepackage{natbib}
\setcitestyle{square,comma,numbers,sort&compress}
\usepackage{graphicx}
\usepackage{subcaption}
\usepackage[dvipsnames]{xcolor}
\usepackage{algorithm}
\usepackage{algpseudocode}
\usepackage{listings}
\usepackage{lineno}
\usepackage{enumitem}
\usepackage{float}
\usepackage[normalem]{ulem}
\usepackage{diagbox}
\usepackage{makecell}
\usepackage{array}
\usepackage{rotating}

\usetikzlibrary{arrows.meta, positioning, calc}

\newcommand{\ours}{TraceForge}                 
\newcommand{\familymodel}{Qwen3.8-Max}         

\RequirePackage{xspace}
\makeatletter
\DeclareRobustCommand\onedot{\futurelet\@let@token\@onedot}
\def\@onedot{\ifx\@let@token.\else.\null\fi\xspace}

\definecolor{darkblue}{rgb}{0, 0, 0.5}
\hypersetup{colorlinks=true, citecolor=darkblue, linkcolor=darkblue, urlcolor=darkblue}

\definecolor{codeblue}{rgb}{0.25, 0.5, 0.5}
\definecolor{codekw}{rgb}{0.35, 0.35, 0.75}
\lstdefinestyle{Pytorch}{
    language = Python,
    backgroundcolor = \color{white},
    basicstyle = \fontsize{9pt}{8pt}\selectfont\ttfamily\bfseries,
    columns = fullflexible,
    aboveskip=1pt,
    belowskip=1pt,
    breaklines = true,
    captionpos = b,
    commentstyle = \color{codeblue},
    keywordstyle = \color{codekw},
}

\input{math_commands}

\makeatother

\newlength\savewidth

\newcolumntype{x}[1]{>{\centering\arraybackslash}p{#1pt}}
\newcolumntype{y}[1]{>{\raggedright\arraybackslash}p{#1pt}}
\newcolumntype{z}[1]{>{\raggedleft\arraybackslash}p{#1pt}}

\makeatletter
\newcommand{\thickhline}{%
 \noalign {\ifnum 0=`}\fi \hrule height 1pt
 \futurelet \reserved@a \@xhline
}
\makeatother

\definecolor{green}{HTML}{009000}
\definecolor{red}{HTML}{ea4335}

\renewcommand{\paragraph}[1]{\vspace{1.25mm}\noindent\textbf{#1}}

\title{\centering \textbf{\ours{}: Distilling with Your Own Thinking}}

\author[1]{Yan Wang}

\affiliation[1]{Independent Researcher}

\email{yanwang.branden@gmail.com}

\input{sections/abstract}

\headercontent{
    \href{https://huggingface.co/datasets/libertywing/TraceForge-Fable-5-Reasoning-Cursor}{Data}\quad
    \href{https://alleai.cloud/cot/}{API}\footnotemark[1]
}

\begin{document}
\thispagestyle{firstheader}
\maketitle
\footnotetext[1]{API supported by ALLE AI Lab. The service currently returns \textsc{Qwen}-family traces only; traces for other model families will follow. Follow the author's RedNote (ID: 9670415842) for updates.}

\definecolor{noticeblue}{RGB}{235, 243, 250}
\definecolor{borderblue}{RGB}{70, 130, 180}
\begin{figure}[H]
    \centering
    \includegraphics[width=0.95\linewidth]{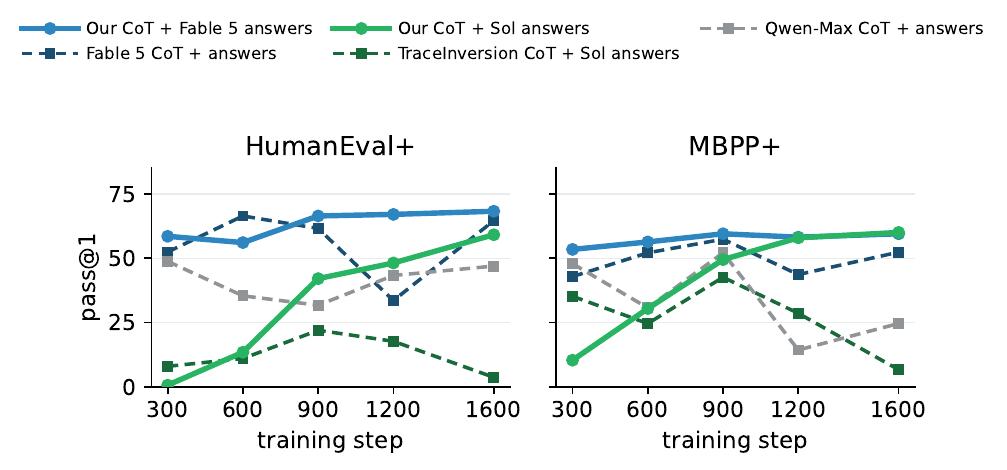}
    \caption{\textbf{Same answers, better traces.} HumanEval+ (left) and MBPP+ (right) across training. Against the model trained on Fable 5's own traces, ours with Fable 5's answers wins at nearly every checkpoint. Both of our models, with Fable 5 and with GPT-6.1-SOL as the teacher, are far better than the one trained with Qwen3.8-Max as teacher.}
    \label{fig:teaser}
\end{figure}

\input{sections/1_introduction}
\input{sections/3_method}
\input{sections/4_experiments}
\input{sections/5_conclusion}
\input{sections/2_related_work}

\bibliographystyle{unsrt}
\bibliography{custom}

\end{document}

%% file: math_commands.tex
\usepackage{amsmath,amsfonts,bm}

\def\1{\bm{1}}

%% file: sections/abstract.tex
\abstract{
Using an LLM to generate training data for another is among the most general ways to train one, and it
remains common practice. Some frontier teams have closed that route for their own models: what comes
back is an answer, or a short summary, without the detailed reasoning trace behind it. That is a little ironic: they were themselves built on distilling human knowledge, and distilling an open-source model is
still entirely routine; the restriction applies to their own outputs alone. Recent work recovers the
hidden reasoning by replaying encrypted blocks across models \cite{panfilov2026stealing}. It somehow
works, but it breaks the providers' terms of service, so it is not something a product can be built on
long-term.

We built the alternative. We take the context and the teacher's answer as input, and generate the
reasoning trace ourselves, in the style of the student's own model family. We call the result a ``family-native'' trace. Training is two-stage,
supervised and then adversarial: a generator learns to regenerate the family's own reasoning traces from their context and answer, and a critic learns how real a reasoning trace looks. Applied to a teacher's answer, the generator writes one in the family's style. An independent judge rates how likely traces are to be genuine on a scale from 0 to 1: ours average $0.68$, real traces $0.66$, and traces selected by the critic from four candidates $0.82$. These are authenticity scores, not correctness rates.

We test the idea on five benchmarks with everything else held fixed. On general tasks the five training corpora are about the same; on coding, the generated reasoning beats Fable~5's own thinking
content. With the same $5{,}000$ questions and the same teacher answers, HumanEval+ improves by $3.7$
points and MBPP+ by $7.1$ at the final checkpoint, and the MBPP+ lead holds at every checkpoint.
We suspect that reasoning written in the student's family style is easier for it to learn from.
}

%% file: sections/1_introduction.tex

\section{Introduction}
\label{sec:Introduction}

Using an LLM to generate training data for another is among the most general ways to train one. A strong
model answers a set of questions, a weaker one is trained on those answers and, better still, on the
reasoning that led to them, and the gap closes \cite{deepseekr1}. This is not a loophole. Training a model on another model's output is how this field has advanced from the start.

Some labs have decided to stop this. The strongest models now return an answer, or a short summary of
their reasoning, and nothing more. The stated reasons are safety and intellectual property; the effect
is a moat, and the moat is the point. Others have tried to go around it by hacking the reasoning traces directly:
recent work shows that the encrypted reasoning blocks returned by providers are interchangeable across
models and can be replayed to recover a hidden reasoning trace verbatim \cite{panfilov2026stealing}. It works, but
it breaks the terms under which those APIs are sold, so it is not something to build on long term.

We introduce TraceForge: keep the teacher's answer, write the reasoning ourselves. We train a generator to regenerate the reasoning traces of the student's own model family from their context and answer, and a critic to keep those reasoning traces indistinguishable from the family's real ones. Applied to a teacher's answer, the generator writes the reasoning that arrives at it, in the family's style. The two are trained against each other, and at inference
the critic picks the best of several candidates, with no labels and nothing that violates a provider's
terms. We call the result a family-native trace.

We train Qwen3.5-9B-Base on reasoning traces produced in different ways. On the two coding benchmarks,
replacing Fable 5's reasoning trace with a family-native one raises pass@1 by $3.7$ and $7.1$ points at the final
checkpoint, and the MBPP+ lead holds at every checkpoint we measured; on the three general-knowledge
tasks, which do not test reasoning, the result is about the same. Reasoning traces synthesized from GPT-6.1-SOL's
answers reach a similar level. Family-native traces also beat training on the family's strongest model directly: with Qwen3.8-Max's own reasoning traces and answers, the same student trails our corpus by $21$ and $35$ points on the same two benchmarks. The teacher's capability sets the ceiling for the student's.

One might be surprised that the generated reasoning traces beat Fable 5's own. We suspect the reason is familiarity: the teacher provides the answer, but the reasoning is written in a style the student can learn from more easily.

Our contribution is easy to state. Training on another model's output is an old and open technique; the only line that matters is whether the data was obtained legally. We use the word
distillation without apology, and we stay on the right side of that line. The community can now learn
from the best models without asking permission and, on the evidence here, end up with a better
student than those models' own reasoning traces would have produced.

%% file: sections/3_method.tex
\section{Method}
\label{sec:method}

\begin{figure}[t]
\centering
\includegraphics[width=\linewidth]{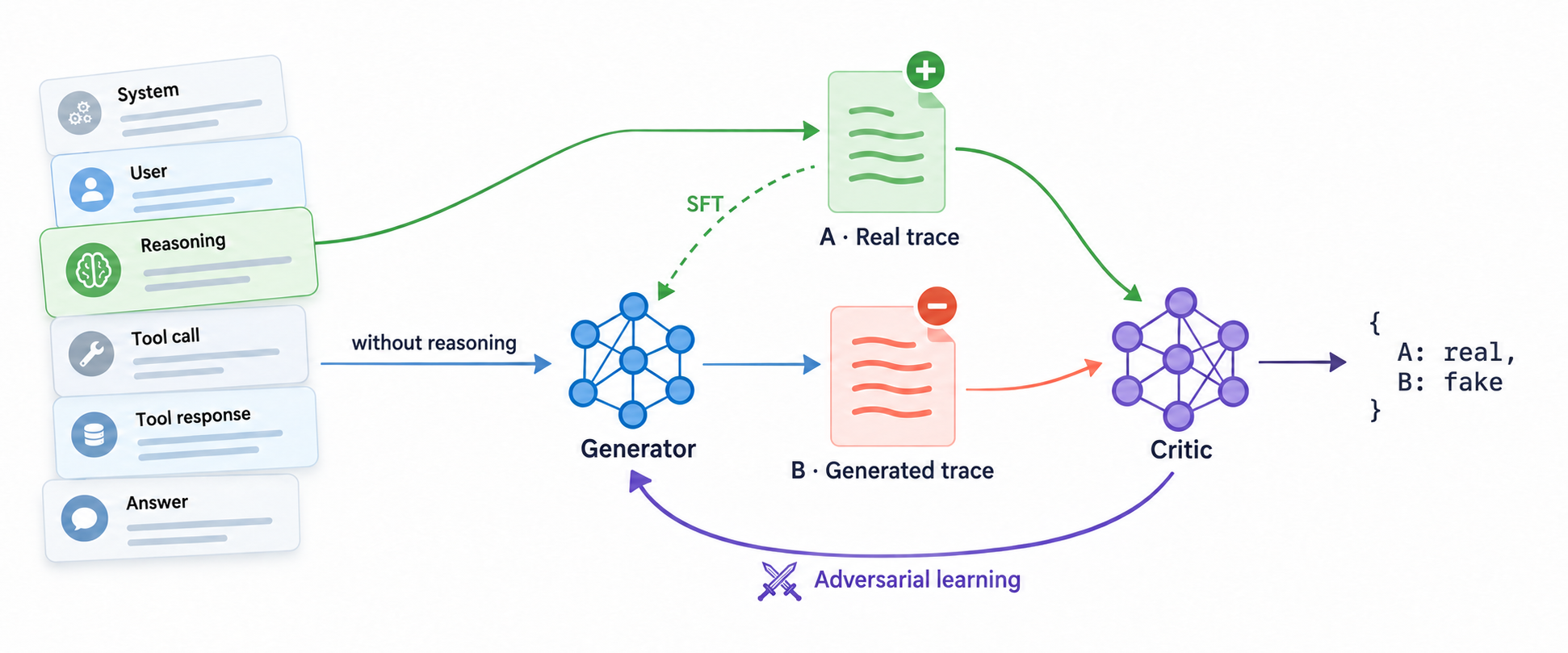}
\caption{\textbf{The pipeline.} A conversation is a system prompt, user turns, tool calls and
responses, an answer, and the reasoning trace that sits before the answer. Training the generator
(blue) takes the conversation with the reasoning trace removed and restores it as the target (green). The critic
(purple) is shown a real reasoning trace and a generated one and must say which is which. The generator is then
trained against the critic's verdict (arc).}
\label{fig:pipeline}
\end{figure}

\subsection{Problem}
\label{subsec:setting}

Figure~\ref{fig:pipeline} shows the object we work with, a conversation written as a message list. Let
$R$ be the reasoning trace and let $S$, $U$, $A$ and $T$ be everything else: the system prompt, the user
turns, the answer, and any tool call or response. The generator models
\[
  P(R \mid S, U, A, T),
\]
the reasoning trace a model would write given the rest of the exchange.

The supervised stages use conversations from the student's own family only: $10{,}000$ examples for the generator (Section~\ref{subsec:generator}) and $10{,}000$ matched real/generated pairs for the critic (Section~\ref{subsec:critic}). In the conversations collected from \familymodel{}, all five elements are present, so $R$ is supervised directly. No outputs from Claude or GPT models are used in these supervised stages. When synthesizing the student's training data, we take $S, U, A, T$ from the teacher, which supplies everything except the reasoning trace, and generate a family-native trace conditioned on those inputs.

\subsection{Supervised training of the generator}
\label{subsec:generator}

The generator is a model from the Qwen family. It is fine-tuned on $10{,}000$ family-model conversations to
maximize $\log P(R \mid S, U, A, T)$: the reasoning trace is removed from the input, everything else is kept, and
the loss falls on the reasoning trace alone. This is the blue path in Figure~\ref{fig:pipeline}. Two details carry
the design. The target is always a reasoning trace the family itself wrote, so the form is learned from data
rather than requested in a prompt. And the answer is part of the input and never part of the output, so
the model learns a derivation that \emph{ends} at $A$ rather than a restatement of it.

One consequence is worth stating, because it is easy to misread the result. The reasoning traces we train on were
written for the family's own answers. When we later ask the generator for a reasoning trace to accompany a
teacher's answer, it cannot be reproducing a reasoning trace it has seen. It is doing what it was trained to do:
writing a derivation that ends at that answer, in the family's voice. We call the result a \emph{family-native} trace. No provider gives us the real
reasoning trace anyway, and none is needed.

After this stage we ask an independent judge, GPT-6.1-SOL, how likely each reasoning trace is to have been written by the family model. We average the assigned probabilities across traces: genuine family traces score $0.66$ and the generator's traces score $0.54$. This measures perceived authenticity, not reasoning correctness. That gap is what the rest of the section addresses.

\subsection{Supervised training of the critic}
\label{subsec:critic}

The critic is trained to tell real reasoning traces from generated ones, the purple path of
Figure~\ref{fig:pipeline}. Its supervised training set contains $10{,}000$ matched pairs: for each genuine family trace, the supervised generator produces one negative. Training stops once the critic assigns the positive a probability at least $0.2$ higher
than that negative, rather than driving the margin as far as it will go.

We made one mistake here that is worth recording. We assumed that a smaller model could judge a larger
one. It cannot: within a few hundred steps the critic fitted shortcuts, scored well on held-out pairs,
and gave the generator nothing to learn from. The critic is therefore at least as large as the
generator.

\subsection{Adversarial co-training}
\label{subsec:loop}

We then run five rounds of online co-training, the arc at the bottom of Figure~\ref{fig:pipeline}. In
each round the generator produces $5{,}000$ reasoning traces, which are shown to the critic alongside the
corresponding real reasoning traces. A reasoning trace that fools the critic becomes critic training data; a reasoning trace that
does not becomes a preference pair for the generator, optimized with the a DPO-like objective~\cite{rafailov2023dpo}. At the end of each round we re-score a held-out set with GPT-6.1-SOL.

The judge's mean authenticity score stops improving after five rounds, and we stop there. At that point the generator's traces score $0.68$, compared with $0.66$ for genuine family traces. Selecting the best of four candidates with the critic raises the generated traces' score to $0.82$, using the same judge and metric defined above.

%% file: sections/4_experiments.tex
\section{Evaluation}
\label{sec:experiments}

The point of this work is to break the moat of the closed-source model providers, so the test that
settles it is a simple one: is a student trained on family-native traces as good as a student trained on
the real ones? Everything in this section follows from that question. We hold the base model, the
training recipe and the budget fixed, vary only where the reasoning trace and the answer come from, and we simply want to see how the two students compare.
\subsection{Setup}
\label{subsec:setup}

\paragraph{Models.} Every student is fine-tuned from
\href{https://huggingface.co/Qwen/Qwen3.5-9B-Base}{\texttt{Qwen/Qwen3.5-9B-Base}}, the pretrained
checkpoint, so that the only difference between two students is their training data.

\paragraph{Data.} We construct $5{,}000$ assistant-turn training examples from Fable~5 coding-agent trajectories released in \href{https://huggingface.co/datasets/TeichAI/Fable-5-Cursor-Traces}{TeichAI/Fable-5-Cursor-Traces}~\cite{teichai2026fablecursor}. We retain conversation contexts up to 256K tokens. Every student-training corpus contains $5{,}000$ examples and uses the same prompts and context; the sources of the reasoning traces and answers vary by setting. S1 and S2 share the original Fable~5 answers, S4 and S5 share newly collected GPT-6.1-SOL answers, and S3 uses \familymodel{}'s own reasoning traces and answers. These student-training corpora are separate from the generator and critic training sets described in Section~\ref{sec:method}.

\begin{itemize}
  \item \textbf{S1, Fable 5 reasoning $+$ answer.} The original data, used as is: for each of the 5{,}000 prompts, Fable 5's real reasoning trace together with its own answer.
  \item \textbf{S2, family-native $+$ Fable 5 answer.} As S1, with the reasoning trace replaced by a
        family-native trace from the generator and critic of Section~\ref{sec:method}; the answers are
        unchanged. S1 and S2 therefore differ only in who wrote the reasoning trace.
  \item \textbf{S3, Qwen-Max reasoning $+$ answer.} The prompts of S1 (all context)  are
        given to \familymodel{}, the strongest legal model in the student's own
        family, and its reasoning traces and answers are used for training.
  \item \textbf{S4, TraceInversion $+$ Sol answer.} We have no GPT-6.1-SOL data carrying real reasoning traces, so
        we mimic S3: we collect GPT-6.1-SOL's answers on the same prompts and
        synthesize a reasoning trace for each with \textsc{TraceInversion} (\textsc{Sol} below), the inversion method of
        Zhang et al.~\cite{zhang2026traceinversion}.
  \item \textbf{S5, family-native $+$ Sol answer.} As S2, applied to S4: our reasoning traces over GPT-6.1-SOL's answers.
        S4 and S5 share the answers and differ only in the synthesizer.
\end{itemize}

\paragraph{Training.} Teacher-forced SFT with the prompt masked from the loss. No truncation: every row
is under 262{,}144 tokens, and packed sequences may reach that length. Full parameters, BF16, AdamW with
weight decay $0.1$ and gradient clipping $1$, learning rate $2\times10^{-5}$ under a cosine schedule
with 5\% warmup down to $2\times10^{-6}$, micro-batch $1$, eight H20 GPUs with TP4/PP2 and sequence
parallelism, seed $42$, no shuffling. One step is one packed-batch update. The five-round adversarial co-training in Section~\ref{subsec:loop} trains the trace generator and critic before synthesis; the downstream students are trained by SFT on the resulting corpora.

\paragraph{Evaluation.} We use three multiple-choice tasks, ARC-Easy, ARC-Challenge and BoolQ, to test
general ability in a non-reasoning mode, and two code benchmarks, HumanEval+ and MBPP+, to test coding
ability in a reasoning mode.
\subsection{Main comparison}
\label{subsec:main}

Table~\ref{tab:main} gives the final checkpoint. The three knowledge benchmarks do not separate the five
corpora: every value sits within $1.6$ points of every other, and no setting leads on all three. That is
what we expect of tasks that never ask for a reasoning trace, and we set them aside here. The code benchmarks are a different picture,
with a spread of more than sixty points.

The first surprise is that S2 beats S1. Replacing the teacher's own reasoning trace with a family-native one adds
$3.7$ points on HumanEval+ and $7.1$ on MBPP+, and Table~\ref{tab:trajectory} shows the MBPP+ lead
holding at every checkpoint. S1 is the textbook recipe, built on the assumption that the strongest
available reasoning trace is the best training signal. On these two benchmarks it is not.

The second surprise is S3. Qwen-Max is a far stronger model than the student, so we expected it to be
almost as good a teacher as Fable 5, perhaps a little behind. It is nowhere near: $18$ points behind on
HumanEval+ and $28$ on MBPP+. A capable teacher writing in its own hand transfers much less than a
stronger teacher whose answers are re-expressed in the student's hand. Teacher capability turns out to
be necessary but not sufficient.

TraceInversion is the weakest of the five, at $3.66$ and $6.61$, and this one did not surprise us once
we read its output. Its reasoning traces are recognisably synthetic: they narrate the solution in a generic voice,
use numbered steps the family never writes, and often restate the answer rather than derive it. The
score matches what a glance at the data already suggests.

S5 closes the argument with a second teacher. Given exactly the same GPT-6.1-SOL answers as S4, our
reasoning traces reach $59.15$ and $60.05$ against $3.66$ and $6.61$: $55$ and $53$ points with nothing varying but
the synthesizer. Set beside S2, which shares the synthesizer and differs only in whose answers it writes,
S5 is $9$ points behind on HumanEval+ and level on MBPP+. That is about the gap one would expect between
the two teachers themselves, and it matches the common view of how the two model families compare.
\begin{table}[t]
\centering
\caption{\textbf{Five corpora at the final checkpoint} ($\approx$1{,}600 packed-batch updates).
Multiple-choice columns are acc-norm for ARC and accuracy for BoolQ; code columns are pass@1 on the
plus suites. S4 reaches its own peak at step 900 ($21.95$ and $42.59$) before falling to $3.66$ and $6.61$ at the final checkpoint. One run per cell.}
\label{tab:main}
\small
\begin{tabular}{p{6.6cm}ccccc}
\toprule
Corpus & ARC-E & ARC-C & BoolQ & HumanEval+ & MBPP+ \\
\midrule
S1: Fable 5 reasoning $+$ answer   & 81.02 & 58.36 & 89.63 & 64.63 & 52.38 \\
S2: family-native $+$ Fable 5 answer & 81.19 & 59.73 & 89.33 & \textbf{68.29} & 59.52 \\
S3: Qwen-Max reasoning $+$ answer  & 81.52 & 58.28 & 88.35 & 46.95 & 24.60 \\
S4: TraceInversion $+$ Sol answer  & 81.48 & 58.53 & 88.99 & 3.66 & 6.61 \\
S5: family-native $+$ Sol answer   & 80.98 & 58.19 & 88.65 & 59.15 & \textbf{60.05} \\
\bottomrule
\end{tabular}
\end{table}
\begin{table}[t]
\centering
\caption{\textbf{Coding accuracy at every checkpoint.} pass@1 on HumanEval+ (upper block) and MBPP+
(lower block); Fig.~\ref{fig:teaser} plots the same numbers.}
\label{tab:trajectory}
\small
\begin{tabular}{lccccc}
\toprule
Step & 300 & 600 & 900 & 1{,}200 & 1{,}600 \\
\midrule
\multicolumn{6}{l}{\emph{HumanEval+}} \\
S1 teacher reasoning     & 52.44 & \textbf{66.46} & 61.59 & 33.54 & 64.63 \\
S2 family-native         & \textbf{58.54} & 56.10 & \textbf{66.46} & \textbf{67.07} & \textbf{68.29} \\
S3 Qwen-Max reasoning    & 48.78 & 35.37 & 31.71 & 43.29 & 46.95 \\
S4 TraceInversion        & 7.93  & 10.98 & 21.95 & 17.68 & 3.66 \\
S5 family-native + Sol   & 0.61  & 13.41 & 42.07 & 48.17 & 59.15 \\
\midrule
\multicolumn{6}{l}{\emph{MBPP+}} \\
S1 teacher reasoning     & 42.86 & 52.12 & 57.41 & 43.65 & 52.38 \\
S2 family-native         & \textbf{53.44} & \textbf{56.35} & \textbf{59.52} & \textbf{58.20} & 59.52 \\
S3 Qwen-Max reasoning    & 47.88 & 30.69 & 52.12 & 14.29 & 24.60 \\
S4 TraceInversion        & 35.19 & 24.60 & 42.59 & 28.57 & 6.61 \\
S5 family-native + Sol   & 10.32 & 30.42 & 49.47 & 57.94 & \textbf{60.05} \\
\bottomrule
\end{tabular}
\end{table}

\subsection{Further analysis}
\label{subsec:further}

We also want to know how each corpus behaves over training, not only where it ends.
Table~\ref{tab:trajectory} reports every intermediate checkpoint on the two code benchmarks, and the
trajectories say more than the endpoints do.

S1 is not stable. It reaches $66.46$ on HumanEval+ at step 600, falls to $33.54$ at step 1{,}200, and
recovers to $64.63$ by the end; we re-ran the 1{,}200-step evaluation twice to rule out a scoring error,
and the number held. The repeated evaluations reproduce this large drop. One possible explanation is that adapting to the teacher's reasoning style temporarily disrupts the student's coding behavior.

S2 and S5 go the other way. S2 shows an overall upward trend with only small interim dips, ending at $68.29$ and $59.52$. S5 rises at every single
checkpoint on both benchmarks, from $0.61$ and $10.32$ to $59.15$ and $60.05$; on the average of the two,
S5 edges past S1.

S5 starts last and finishes near the front. Like S2, it uses family-native reasoning traces synthesized by the same generator-and-critic pipeline, but conditioned on a different teacher's answers. This comparison therefore changes the answer source while keeping the trace-synthesis procedure fixed. Qwen3.5's mid-training is widely said to have used a lot
of Claude-family data and very little GPT-family data, which would make GPT-6.1-SOL's answer style new to
the base model: it needs steps to adapt to it before the reasoning trace can help. The reasoning trace style is not new, which
is why the climb, once it starts, never reverses.

We also have some interesting findings on S3 and S4. Both had their moment and both ended badly: S3
reaches $48.78$ on HumanEval+ inside the first 300 steps and never gets back, and S4 peaks at $21.95$ and
$42.59$ at step 900 before falling to $3.66$ and $6.61$. Inspecting their outputs, we observe increasing dead loops and meaningless repetition: sentences recur with small variations, or the model repeatedly re-derives an existing result. This degeneration accompanies the performance decline and may contribute to it.

This matches a piece of commonsense among foundation-model teams. Distilling from a distilled model, which people call
``breathing second-hand smoke'', may result in a student that tends to fall into dead loops while it thinks;
in contrast, train on Claude or GPT directly and the problem rarely appears. Intuitively, we guess what happened to S3 and S4 is, in some measure,
explained by that one habit. S1, which takes its reasoning traces and answers from Fable 5 directly, never shows the pattern. Finally, we were pleased to find that our family-native traces do not produce it at all.

The two tables together settle the comparison. A family-native trace is not merely a legal
substitute for a real one. On the comparison that isolates the reasoning trace, with questions and answers held
fixed, \textbf{it is even the better teacher}.

%% file: sections/5_conclusion.tex
\section{Discussion and Conclusion}
\label{sec:discussion}

The result is easy to state. Our reasoning traces beat the strongest model we are allowed to distil from,
and they beat it by a wide margin: against Qwen-Max's own reasoning traces and answers we are $21$ and $35$
points ahead. They also edge past the teacher's own reasoning traces, by $3.7$ and $7.1$ points with the
questions and answers held fixed. A comparison from schooling fits. The best student is not produced by the best teacher alone, but by that teacher teaching in the way the student learns best. Our traces are
that lesson plan: the strongest teacher's content, delivered in a way that students are familiar with.

Neither the generator nor the critic has ever seen a Claude or an OpenAI token. Every training example comes from Qwen3.8-Max, a model whose terms permit this
use, so the pipeline violates no provider's agreement. 

One more option, for teams that still see risk in handling Claude or OpenAI answers at all. Generate the
family-native trace, hand it to the strongest legally-distillable model in the same family such as Qwen3.8-Max, and let that
model write the answer. Then train on the family-native trace and that answer together. No Claude or
OpenAI token is backpropagated at any point, which removes both the legal exposure and the provenance
question: \textbf{the pipeline yields data that is, on every count, the family's own}. We have run this variant, and it performs essentially the same as S2.

%% file: sections/2_related_work.tex
\section{Related Work}
\label{sec:related}

\paragraph{Learning from reasoning traces.} Reasoning traces provide supervision beyond final answers.
Distilling Step-by-Step uses teacher-generated rationales as an additional training target for smaller
task-specific models~\cite{hsieh2023distilling}; Orca learns from explanation traces generated by larger
models~\cite{mukherjee2023orca}; and DeepSeek-R1 transfers reasoning capabilities to smaller models through
curated reasoning data~\cite{deepseekr1}. These works motivate training on intermediate reasoning. Our
question concerns its source and form: when a teacher's answer is fixed, can a trace synthesized in the
student's model-family style teach more effectively than the teacher's original trace?

\paragraph{Recovering and synthesizing hidden reasoning.} Panfilov et al.~\cite{panfilov2026stealing}
recover protected reasoning by transferring encrypted blocks across models. Synthesis takes a different
route: Trace Inversion learns to generate reasoning from inputs, answers and optional
summaries~\cite{zhang2026traceinversion}, while REER works backwards from known-good solutions for
open-ended generation~\cite{wang2026reer}. Trace Inversion already shows that synthesized traces can
outperform answer-only training and, in some settings, original traces; it also reports weaker coding
results associated with style and format mismatch. TraceForge instead trains its generator on the
student family's own traces and uses adversarial feedback from a family-trace critic. Our S1/S2
comparison holds Fable~5's answers fixed and replaces only the trace source; S4/S5 holds GPT-6.1-SOL's
answers fixed and compares the two synthesis pipelines. These comparisons establish downstream
performance differences, but do not isolate family style from every other difference between the pipelines.

\paragraph{Teacher--student distribution differences.} MiniLLM changes the distillation objective to
reverse KL to better account for the student's approximation of a generative teacher
distribution~\cite{gu2023minillm}. Our intervention is in the training text: we retain an answer from a
strong teacher while synthesizing its accompanying reasoning in a family-native form. This separates
the answer source from the trace source. S3, which uses \familymodel{}'s own answers and traces, is a
distinct baseline; its performance should not be read as an ablation of trace style with answers held fixed.

\paragraph{Selection, verification and rewriting.} STaR bootstraps reasoning by retaining successful
solutions and using correct answers to guide further rationale generation~\cite{zelikman2022star}.
Process supervision trains verifiers with step-level correctness labels~\cite{lightman2023verify}, and
self-consistency aggregates sampled reasoning paths by answer agreement~\cite{wang2022selfconsistency}.
Our critic instead learns to distinguish genuine family traces from generated ones. It supports both
generator training and best-of-four selection without step-level correctness labels. Its authenticity
score and the independent judge's score measure resemblance to family traces; downstream benchmark
performance is evaluated separately.

\paragraph{Adversarial training and the reconstruction target.} Self-play fine-tuning contrasts model
outputs with human demonstrations~\cite{chen2024spin}, and DPO learns from preference
pairs~\cite{rafailov2023dpo}. We use a family-trace critic to construct online preference data for the
generator. Instruction backtranslation reconstructs an instruction from a response~\cite{li2023humpback};
our reconstruction target is the reasoning trace conditioned on the conversation and answer. Finally,
written reasoning need not faithfully reveal the computation that produced an
answer~\cite{turpin2023unfaithful}. Our goal is a useful family-native training trace, rather than a
verbatim reconstruction of the teacher's hidden reasoning.